\documentclass[runningheads]{llncs}
\usepackage{graphicx}
\usepackage{amsfonts}
\usepackage{booktabs}
\usepackage{amsmath}
\usepackage{multirow}
\begin{document}
\title{Region Based Adversarial Synthesis of Facial Action Units}
%
%
%
%
%
\author{Zhilei Liu\thanks{Corresponding author.} \and Diyi Liu \and Yunpeng Wu}
\institute{College of Intelligence and Computing, Tianjin University \\
\email{\{zhileiliu, liudiyi, wuyupeng\}@tju.edu.cn} }

\maketitle              
\begin{abstract}
Facial expression synthesis or editing has recently received increasing attention in the field of affective computing and facial expression modeling. However, most existing facial expression synthesis works are limited in paired training data, low resolution, identity information damaging, and so on. To address those limitations, this paper introduces a novel Action Unit (AU) level facial expression synthesis method called Local Attentive Conditional Generative Adversarial Network (LAC-GAN) based on face action units annotations. Given desired AU labels, LAC-GAN utilizes local AU regional rules to control the status of each AU and attentive mechanism to combine several of them into the whole photo-realistic facial expressions or arbitrary facial expressions. In addition, unpaired training data is utilized in our proposed method to train the manipulation module with the corresponding AU labels, which learns a mapping between a facial expression manifold. Extensive qualitative and quantitative evaluations are conducted on commonly used BP4D dataset to verify the effectiveness of our proposed AU synthesis method.

\keywords{Facial action unit  \and Facial expression synthesis/editing \and Conditional generative adversarial network.}
\end{abstract}

\vspace{-1mm}
\section{Introduction}
\vspace{-1mm}
Recently, identity preserving facial expression generation or synthesis from a single facial image has attracted continuously increasing attention in the field of affective computing and computer vision~\cite{zhou2017CDAAE,GANimation2018,Ding2017ExprGAN}. Ekman and Friesen \cite{ekman1997FACS} developed the Facial Action Coding System (FACS) for defining facial expressions with some basic facial action units (AUs), each of which represents a basic facial muscle movement or expression change. However, with the challenges of time-consuming AU annotation and imbalanced databases, FACS based facial expression analysis with large-scaled data dependent deep learning methods is not widely conducted. To address those issues, facial expression synthesis combined and coordinated by a certain association of AUs has recently received increased attention in the method of facial editing and transformation~\cite{zhou2017CDAAE,GANimation2018}. The generated facial expressions corresponding to desired AU labels can be applied to create a facial expression dataset with multiple diverse AU labels.\par

Most recent existing studies working on facial expression synthesis based on distinct AU states pay more attention on the global face and some of them train with the paired data. Zhou et al.~\cite{zhou2017CDAAE} proposed a conditional difference adversarial autoencoder (CDAAE) to transfer AUs from absence to presence on the global face. This method divides the database into two groups according to absence or presence of AU labels and trains with paired images of the same subject, which increases the complexity of training and the difficulties of preprocess. Moreover, CDAAE employs the low resolution images, which can lose facial details vital for AU synthesis. GANimation proposed by Pumarola et al.~\cite{GANimation2018}, transfers AUs on the whole face which can produce a co-generated phenomenon between different AUs. It is difficult for the method to synthesize a single AU respectively without keeping the other AU untouched. Liu et al.~\cite{Liu2018_3Dmodel} uses 3D Morphable Model (3DMM) to achieve AU synthesis, where the transformation from a image to 3D model damages the texture details of the original images.\par

In this paper, aiming at building a model for facial action unit synthesis with more local texture details, we propose a facial action unit synthesis framework named LAC-GAN by integrating local AU regions with conditional generative adversarial network. With personal identity information well preserved, our proposed LAC-GAN manipulates AUs between different states, which learns a mapping between a facial manifold related to AU manipulation same as CAAE proposed by Zhang et al.~\cite{zhang2017CAAE}. Specifically, our AU manipulation model is trained on unpaired samples and the corresponding AU labels without grouping the database by different states of labels. In addition, the key point of our method is to make the manipulation module focus only on the synthesis of local AU region without touching the rest identity information and the other AUs. For this purpose, AU region of interest(ROI) localization  mechanism without co-generated phenomenon between different AUs is applied to synthesize a single AU respectively and correctly. Finally, to evaluate the performance of proposed method, quantitative analyses on our synthesized facial images are conducted with state-of-the-art facial action unit detectors.

\par
In summary, the contributions of our work are as follows: (1) The proposed AU manipulation module learns a mapping between the facial expression manifold with unpaired samples and corresponding AU labels. (2) Specific AU related local attention is considered to manipulate AU features locally. (3) The proposed AU synthesis framework LAC-GAN enables facial expression synthesis given any desired AU combination of AU combinations. To demonstrate the effectiveness of our proposed framework, both qualitative and quantitative evaluations are performed.

\vspace{-1mm}
\section{Related Work}
\vspace{-1mm}
Our proposed framework is closely related to existing studies on GAN based image generation and facial expression synthesis.

\vspace{-1mm}
\subsection{Conditional Generative Adversarial Network}
\vspace{-1mm}
Generative Adversarial Network (GAN)~\cite{Goodfellow2014GAN} and Deep Convoluntional GAN (DCGAN)~\cite{Radford2016DCGAN} are powerful class of generative models based on game theory. The typical GAN optimization consists in simultaneously training a generator network to produce realistic fake samples and a discriminator network to distinguish between real and fake data. This idea is embedded by the so-called adversarial loss.\par

An active area of research is designing CGAN\cite{Mirza2014CGAN} models extended by GAN that incorporate conditions and constraints into the generation process. Prior studies have explored combining several conditions on transformation task, such as class information~\cite{Odena2017ACGAN}, particularly, interesting for which are those methods exploring image based conditioning as in image super-resolution ~\cite{Ledig2017SRGAN}, image in-painting ~\cite{Pathak2016inpainting}, image-to-image translation ~\cite{Isola2017Translation} and face age progression or regression~\cite{zhang2017CAAE,Wang2018faceaging}.Facial expression editing mostly focus on using the methods of CGAN, similar to the above transferring work.Based on face age progression method~\cite{zhang2017CAAE}, we proposed a conditional model to learn the face manifold conditioned on AUs. 

\vspace{-1mm}
\subsection{Facial Expression Synthesis}
\vspace{-1mm}
In recent years, the study on facial expression editing/synthesis has been actively investigated in computer vision. Recently, Yeh et al.~\cite{Yeh2016SemanticExpr} proposed to edit the facial expression by image warping with appearance flow. Although the model can generate high-resolution images, paired samples as well as the labeled query image are required. GANimation~\cite{GANimation2018} proposed a method of virtual avatar animation on the whole facial images while training a cycle conditional generative adversarial network to achieve the transformation of expression. Recently, Zhou et al.~\cite{zhou2017CDAAE} proposed a conditional difference adversarial autoencoder (CDAAE) for facial expression synthesis that considers AU labels. However, the resolution of its generated facial image is only $32 \times 32$, and the generated facial images with AU labels are not well quantitatively evaluated. By integrating with the 3D Morphable Model (3DMM), Liu et al.~\cite{Liu2018_3Dmodel} proposed an 3D AU synthesis framework to transfer AUs in the range of intensities, however, some texture details are lost for certain AUs because of the limitations of 3D face model. In addition, while generating a single AU, all these above mentioned methods can damage the other AU without keeping the identity information untouched. \par
It is a certain reference for AU synthesis that most works on AU detection make full use of landmark-based geometry to improve the performance.  Li et al.~\cite{Li2017EAC-Net} proposed a deep learning based approach named EAC-Net for facial action unit detection by enhancing and cropping the AU regions of interest(ROI) with roughly extracted facial landmark information.Therefore, in this paper, imitating EAC-Net, our proposed framework separates the facial image into multiple local AU regions which are integrated in modern deep learning model.In addition, attention-GAN ~\cite{Chen2018AttGAN} proposed adopts attention mechanism to focus on generating objects of interests without touching the background region. Similar to them, we proposed AU ROI localization module to maintain the other information except the 
concerned AU.

\vspace{-2mm}
\section{ LAC-GAN for Facial Action Unit Synthesis}
\begin{figure*}[!htbp] 
\centering 
\includegraphics[height=6.75cm, width=12cm]{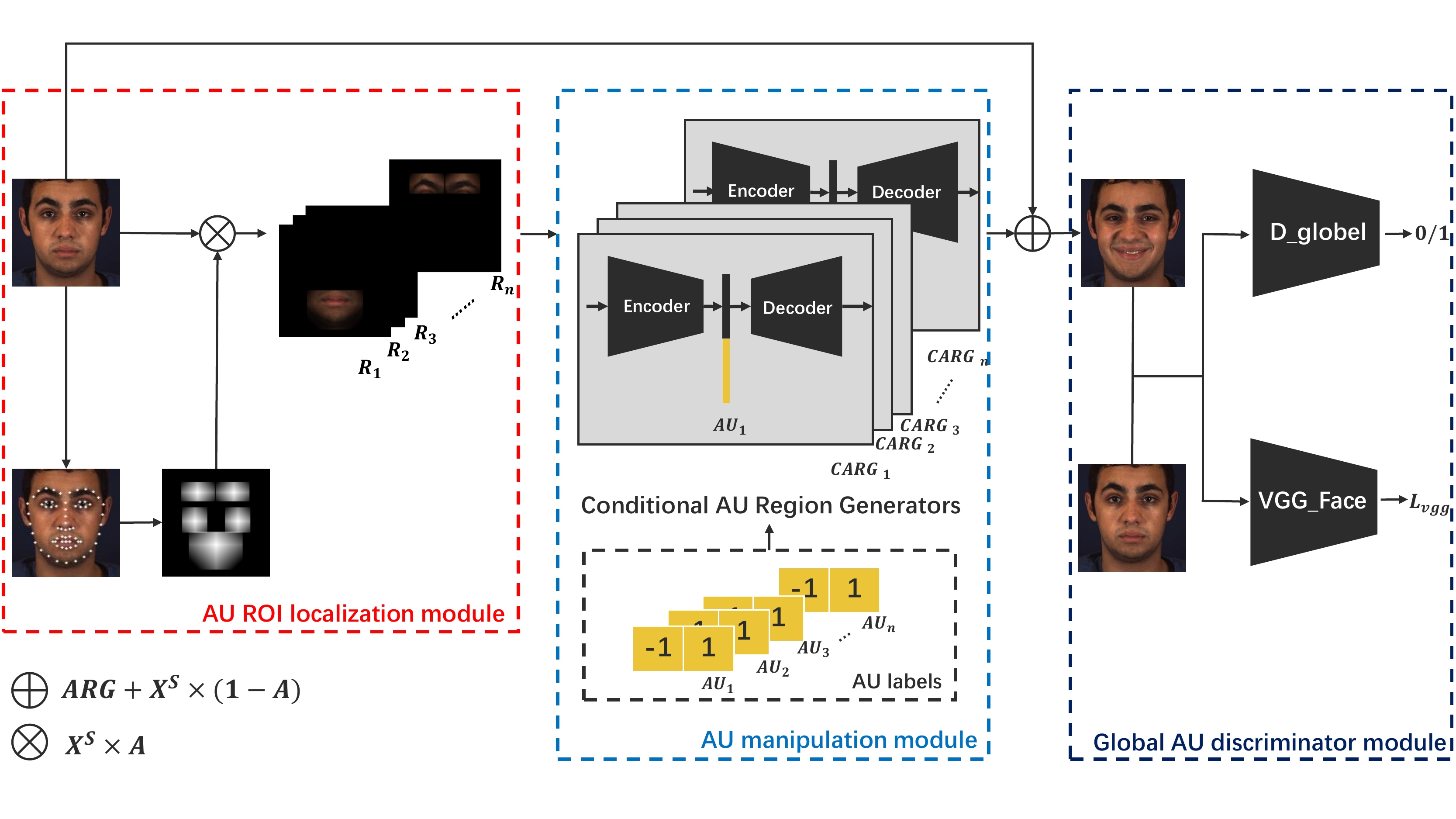} 
\caption{The Overall Framework of LAC-GAN}\label{fig1} 
\end{figure*}

\vspace{-2mm}
In this section, the overall framework of our proposed LAC-GAN for AU synthesis as shown in Fig.~\ref{fig1} is presented at first. The details of the AU manipulation module with conditional AU region generator (CARG) for single AU synthesis as shown in Fig.~\ref{fig2} are introduced in the next.

\vspace{-2mm}
\vspace{-2mm}

\vspace{-6mm}
\subsection{The Overall Framework of LAC-GAN}
\vspace{-1mm}

AS shown in Fig.~\ref{fig1}, our proposed LAC-GAN consists of three modules: AU ROI localization module with fixed local attention, AU manipulation module, and global AU discriminator module. In the AU ROI localization module, AU specific ROIs are defined as local regions of the original facial image $X^s \in \mathbb{R}^ {H\times{W}\times 3}$. Given the input image $X^s$ and the target one-hot AU label vector $l_{AU_i} \in \left\{ -1,+1\right\}^{N_{AU}\times{2}}$, the AU manipulation module focuses on transforming the AU ROI from the source state to the target state. The resulting image is therefore a combination of all the changed AU regions with the rest regions of original facial image. Finally, the global AU discriminator aims to distinguish the real images and the generated images.

\vspace{-1mm}
\subsubsection{AU ROI localization module:} AUs appear in specific facial regions that are not limited to facial landmark points. Instead of directly using facial landmarks, AU center rules are  adopted by us to localize AU specific ROIs. Similar to EAC-net, AU local attention map containing several AUs are designed here for the purpose of generating AU ROIs. In addition, we capture the AU local attention map by applying the Manhattan distance to the AU center. The local AU ROI is defined as:

\vspace{-1mm}
\begin{equation}\label{equ1}
X^{AU_i} = A_{AU_i}\times{X^s},
\end{equation}
\vspace{-1mm}

where $X^{AU_i}$ is the pixel-level weight of the $AU_i$ with the local attention map $A_{AU_i}\in \left\{ 0,...,1\right\}^{H\times{W}}$. 

In order to preserve the original facial information outside of the AU ROIs, an local AU attention localization is considered in this module. Concretely, the whole face is divided into multiple AU regions based on the local AU attention mapping rules. After obtaining AU ROIs, the desired AU regions with target AU labels will be generated by CARGs introduced in section~\ref{Sec4.2}. The generated target image can be obtained as:

\vspace{-1mm}
\begin{equation}\label{equ2}
X^t=\sum_{i=1}^{N_{AU}}{G_{AU_i}(X^{AU_i},l_{AU_i})}+(1-\sum_{i=1}^{N_{AU}}A_{AU_i})\times{X^s},
\end{equation}
\vspace{-1mm}

where $G_{AU_i}(\cdot)$ is the output of the generator $G$ conditioned on target AU label $l_{AU_i}$ and the input ROI $X^{AU_i}$ of $AU_i$ computed in Eq.~\ref{equ1}. 

\vspace{-1mm}
\subsubsection{Global AU Discriminator Module:} In order to improve the photo-realism of the generated images, an adversarial loss between the generated image $X^t$ and the real image $X^r$ is defined as:

\vspace{-1mm}
\begin{equation}\label{Eq_adv}
\min\limits_{G_{AU}}\max \limits_{D_{img}}L^{img}_{adv} =\mathbb{E}_{{X}\sim{P_{data}}(X)}[\log D_{img}(X^r)]+\mathbb {E}_{{X}\sim{P_{data}}(X)}[\log (1-D_{img}(X^t))],  
\end{equation}
\vspace{-1mm}

where $X^t$ is synthetic output obtained by Equ.~\ref{equ2}.

To preserve more identity information between $X^t$ and $X^r$, a pre-trained VGG-face model is leveraged to enforce the similarity in the feature space with following loss:

\vspace{-1mm}
\begin{equation}
\min\limits_{G_{AU}} L_{id} = \sum_l\alpha_lL_1(\phi_l(X^t),\phi_l(X^r)),
\end{equation}
\vspace{-1mm}

where $\phi_l(\cdot)$ is the feature map of the $l_{th}$ layer of the VGG-face network, and $\alpha_l$ is the corresponding weight. Similar to ~\cite{Wang2015DeepFace}, the activations at the $conv1\_2$, $conv2\_2$, $conv3\_2$, $conv4\_2$ and $conv5\_2$ layer of the VGG-face model are used.

\vspace{-2mm}
\begin{figure}[htbp] 
\centering 
\includegraphics[height=6.5 cm, width=12cm]{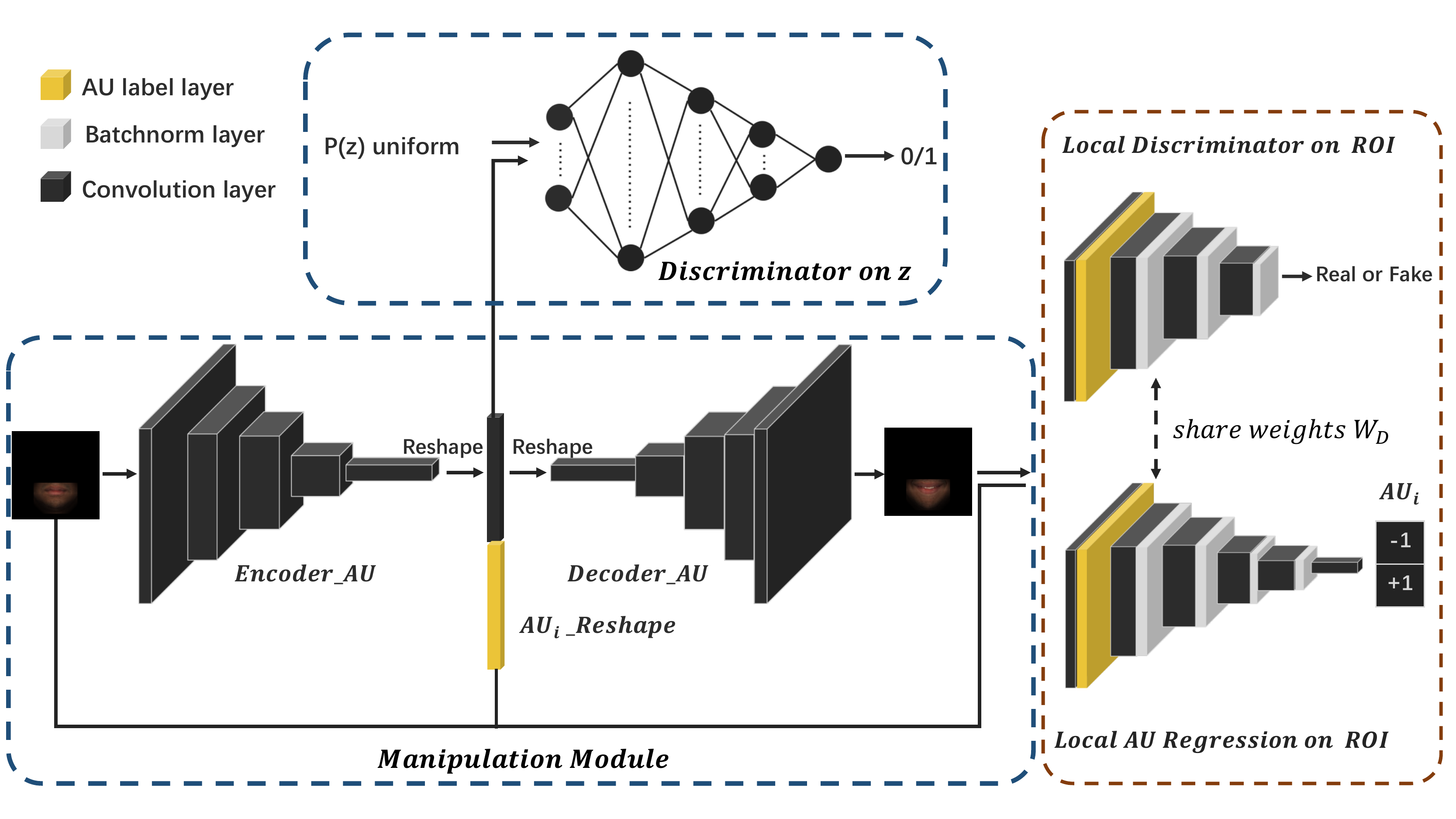} 
\caption{Proposed Conditional AU Regional Generator of AU Manipulation Module}\label{fig2} 
\end{figure}
\vspace{-2mm}

\vspace{-1mm}
\subsection{AU Manipulation Module:}\label{Sec4.2}
\vspace{-1mm}
As shown in Fig.~\ref{fig2}, the proposed AU manipulation module contains multiple local conditional AU regional generators (CARGs) for single facial action unit synthesis. Base on the face manifold assumption demonstrated experimentally in CAAE, the inputs of our proposed CARG are the source ROI of $X^{AU_i}$ and the target label $l_{AU_i}$. On the other hand, additional local critic discriminator is used to evaluate the quality of the generated regional image.

\vspace{-1mm}
\subsubsection{Local Conditional AU Regional Generators:} Given the specific AU ROI $X^{AU_i}$ of the source image and the desired AU label, the generator network $G_{AU_i} = (G^{AU_i}_{en}, G^{AU_i}_{de})$ with encoder-decoder structure is proposed. At first, the encoder $G^{AU_i}_{en}$ maps the input $X^{AU_i}$ into a latent feature vector $z_{AU_i}$, ${N_z}$ is the dimension of the latent feature vector $z_{AU_i}$. By concatenating the obtained $z_{AU_i}$ and the desired AU labels $l_{AU_i}$ together, the decoder $G^{AU_i}_{de}$ is designed to generate the target AU ROI $\hat{X^{AU_i}} = G^{AU_i}_{de}(G^{AU_i}_{en}(X^{AU_i}),l_{AU_i})$ with original identity information and target AU information. A pixel-wise image reconstruction loss is defined as:

\vspace{-1mm}
\begin{equation}
\min\limits_{G^{AU_i}_{en}, G^{AU_i}_{de}}L_{pixel}^i =\mathbb{E}_{{X^{AU_i}}\sim{P_{data}}(X^{AU_i})}[\Vert{G^{AU_i}_{de}(G^{AU_i}_{en}(X^{AU_i}),l_{AU_i})-{X^{AU_i}}}\Vert_1],
\end{equation}
\vspace{-1mm}

in which, $l_1$-norm is adopted to capture low-level structure information. In preliminary experiments, we also tried replacing $l_1$-norm with the other perceptual loss, although we did not observe significant improved performance.

\vspace{-1mm}
\subsubsection{Discriminator $D_z$ on Latent Feature:} The fact is demonstrably authenticated that face images lie on a manifold \cite{He2005laplacianfaces,Lee2003manifolds}. For the purpose of maintaining the AU regions generated by the face manifold \cite{zhang2017CAAE}, we employ the uniform distribution to $z_{AU_i}$ through the discriminator $D_z^{AU_i}$ ,which imposes $z_{AU_i}$ on the uniform distribution without "holes". The adversarial training process is of $D_z$ is defined as:

\vspace{-1mm}
\begin{equation}
\begin{split}
\min\limits_{G^{AU_i}_{en}}\max \limits_{D_z^{AU_i}}L_{adv\_z}^i =& \mathbb{E}_{{z}\sim{P(z)}}[\log{D_z^{AU_i}}(z)]+\\
&\mathbb{E}_{{X^{AU_i}}\sim{P_{data}}(X^{AU_i})}[\log(1-D_z^{AU_i}(G^{AU_i}_{en}(X^{AU_i}))].     
\end{split}
\end{equation}
\vspace{-1mm}

\vspace{-1mm}
\subsubsection{Local AU Discriminator:}  Similar to recent studies conducted by Huang et al.~\cite{Huang2017faceRotation} and  Liu et al. \cite{TranD2014Disface}, an regional adversarial loss between the real AU ROI ${X^{AU_i}}$ and the generated AU ROI $\hat{X^{AU_i}} = G^{AU_i}_{de}(G^{AU_i}_{en}(X^{AU_i}),l_{AU_i})$ is defined. While reducing the global image adversarial loss, the local AU discriminator must also reduce the error, which is defined with two components: a local AU adversarial loss is used to 
distinguish the real and fake AU regions which learns to render realistic samples; and an AUs regression loss of AU regions is used to learn the regression layer on top of local discriminator, which satisfies the target facial expression encoded by $l_{AU_i}$. Those loss can therefore be defined as:

\vspace{-1mm}
\begin{equation}
\begin{split}
\min\limits_{G^{AU_i}_{en}, G^{AU_i}_{de},{R_{AU_i}}}\max \limits_{D_{AU_i}}L_{adv\_AU}^i =& \mathbb{E}_{{X^{AU_i},l_{AU_i}}\sim{P_{data}}(X,l)}[\log{D_{AU_i}}(X^{AU_i},l_{AU_i})]+\\
&\mathbb{E}_{{X^{AU_i},l_{AU_i}}\sim{P_{data}}(X,l)}[\log(1-{{D_{AU_i}}(\hat{X^{AU_i}},l_{AU_i})})]\\
&+\lambda_{AU}L_{label}^i,
\end{split}
\end{equation}
\vspace{-1mm}

where $\lambda_{AU}L_{label}$ is a trade-off parameter of the AU regression loss which keeps the facial region with a specific AU state generated by manipulating the corresponding expression code. This loss is completed as follow:

\vspace{-1mm}
\begin{equation}
\min\limits_{R_{AU_i}}L_{label}^i = \mathbb{E}_{{X^{AU_i},l_{AU_i}}\sim{P_{data}}(X,l)}[\Vert{R_{AU_i}(\hat{X^{AU_i}},l_{AU_i})-l_{AU_i}}\Vert^2_2]
\end{equation}
\vspace{-1mm}

\vspace{-1mm}
\subsubsection{Overall Objective Function:} To generate the target face with desired AU label vector, following loss function $L_{CARG}$ is defined by linearly combining all previous partial losses:

\vspace{-1mm}
\begin{equation}\label{equ10}
\min\limits_{G^{AU_i}_{en}, G^{AU_i}_{de},{R_{AU_i}}}\max \limits_{{D_{AU_i}},{D_z^{AU_i}}}L_{CARG}^i = L_{pixel}^i + {\lambda_1}L_{adv\_z}^i + {\lambda_2}L_{adv\_AU}^i .
\end{equation}
\vspace{-1mm}

The final training loss function of LAC-GAN is a weighted sum of all these losses defined above:

\vspace{-1mm}
\begin{equation}\label{equ11}
\min\limits_{G_{AU},{R_{AU}}}\max \limits_{{D_{AU}},{D_z^{AU}}}L_{LAC-GAN} = \sum_{i=1}^{N_{AU}}\beta_iL_{CARG}^i + {\lambda_3}L_{id}+{\lambda_4}L_{tv}.
\end{equation}
\vspace{-1mm}

where ${\lambda_j}(j = 1,2,3,4)$ is a trade-off parameter,${\beta_i}(j = 1,...,N_{AU}) $ is the corresponding weight of $CARG_i$ loss.The total variation regularization $L_{tv}$ \cite{Mahendran2015tvloss}is adopt on the reconstructed image to reduce spike artifacts.

\vspace{-2mm}
\section{Experimental Evaluation}
\vspace{-2mm}

\vspace{-1mm}
\subsection{Experimental Dataset and Implementation Details}
\vspace{-1mm}
\subsubsection{Dataset:} Our LAC-GAN is evaluated on widely used database for facial AU detection, named BP4D\cite{BP4D2014}. BP4D contains 41 participants with 23 female and 18 male, each of which is involved in 8 sessions captured both 2D and 3D videos for each participant. Each video frame is manually coded with an intensity for each of the 12 AUs, namely AU1, AU2, AU4, AU6, AU7, AU10, AU12, AU14, AU15, AU17, AU23, and AU24, with AU labels of occurrence or absence according to the FACS. In this work, all annotated video frames with successful face registration (100,767 frames) of 31 subjects are selected as the training set, and all annotated video frames (39,233 frames) for the remaining 10 subjects are used as the testing set.The data partition rule is the same as the AU detection model JAA-Net ~\cite{Shao2018JAA} with two folds for training and the remaining one for testing.

\vspace{-2mm}
\subsubsection{Implementation Details:} For each facial image, similarity transformations including rotation, uniform scaling, translation, and normalization are performed to obtain a $200\times200\times3$ color facial image.The kernel size of the multiple convolution and de-convolution layers in our proposed framework is set to be $5\times5$. The encoder consists of 5 convolutional layers and a reshaped fully connection layer, while the decoder consists of the other reshaped layer and 5 de-convolutional layers. The dimensions of $z$ and AU labels are set to be 60. Discriminator $D_z$ employs four fully connection layers. The batch normalization is adopted on local AU discriminator to make the framework more stable. Simultaneously. local AU Regression shares the weight of 4 convolution layers with local AU discriminator.\par

All models are trained using Tensorflow using the Adam optimizer \cite{Kingma2014tvAdam}, with learning rate of 0.0002, $\alpha_1 = 0.5, \alpha_2 = 0.9, \alpha_3 = 0.75, \alpha_4 = \alpha_5 =1$. The weight coefficients for the loss terms in Eq.~\ref{equ10} and Eq.~\ref{equ11} are set to $\lambda_1= 0.01, \lambda_2= 0.1, \lambda_3= 1, \lambda_4= 0.001, \lambda_{AU}= 10$. During training, we set $\beta_i= 1$, where $i$ is corresponded with original AU labels.

\vspace{-2mm}
\subsection{Qualitative Analysis of AU Synthesis Performance }

\subsubsection{Synthetic Results of Single AU:} 
\begin{figure*}[!htb]
\centering
\includegraphics[width=\textwidth]{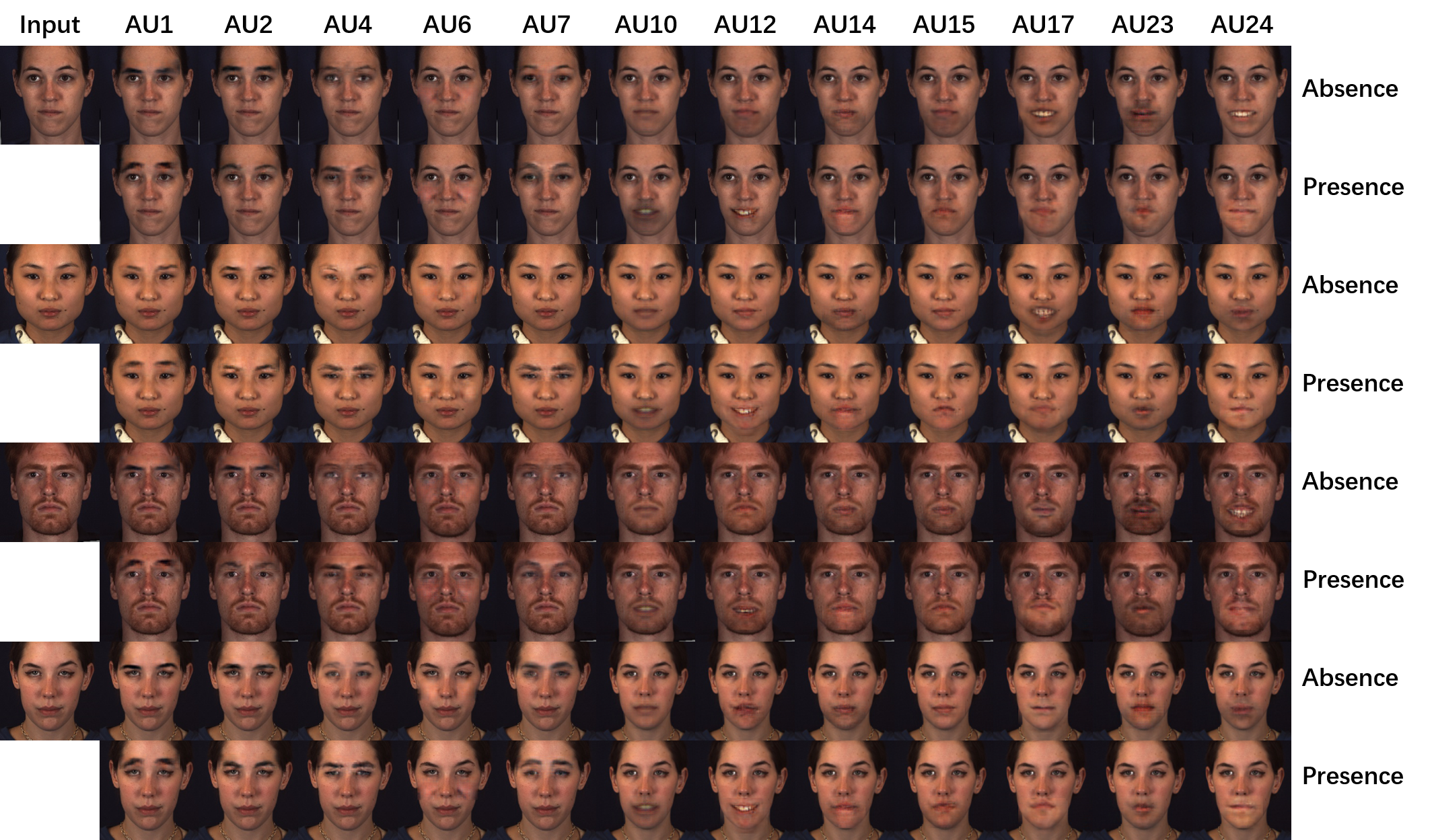}
\caption{Synthetic Results of Single AU.} \label{fig3}
\end{figure*}
\vspace{-2mm}
We firstly evaluate our model's ability to activate the status of AUs by transforming the neutral face to the absence or the presence of specific AU while preserving the person's identity. Fig.~\ref{fig3} shows a subset of 12 AUs individually manipulated at different status during the testing session.\par
For AU1 (inner brow raiser), AU2 (outer brow raiser), and AU4 (inner brow lowerer) located in the brow region, the changes of synthesis face between absence and presence of AU can be obviously perceived. For AU6 (cheek Raiser) located in the cheek region, the movement of cheek muscle can be observed. For AU10 (Upper Lip Raiser), AU12 (Lip Corner Puller), AU14 (Dimpler), and AU15 (Lip Corner Depressor) located in the mouth region, obvious movements of AU muscles can be observed in the synthetic images. For the other AUs corresponding to the movements in the mouth and chin regions, AU23 (Lip Tightener) can be observed like pouting and AU24 (Lip Pressor) seem like closing mouse obviously. However, Subtle variation of AU7 (Lid Tightener) and AU17 (Chin Raiser) which accounts for minor wrinkling in the synthetic faces.

\vspace{-5mm}

\begin{figure}[!htb]
\centering
\includegraphics[width=\textwidth]{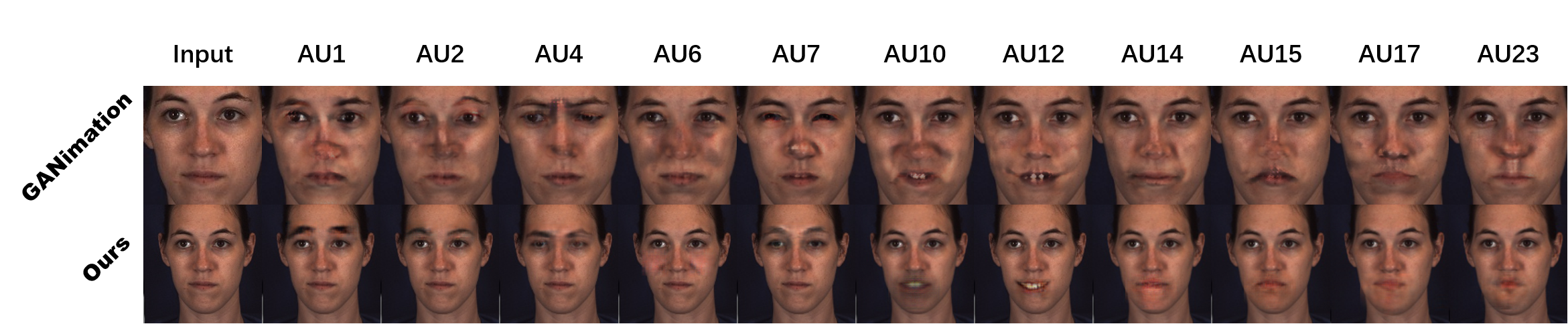}
\caption{Qualitative comparison with GANimation~\cite{GANimation2018}.} \label{fig4} 
\end{figure}
\vspace{-2mm}

\vspace{-6mm}
\subsubsection{Qualitative comparison with GANimation:}
We compare our approach again the baseline GANimation ~\cite{GANimation2018}. For a fair comparison, we adopt the same dataset and experimental details with training GANimation.Fig.~\ref{fig4} shows our method performs better than GANimation on synthetic of single AUs, with keeping identity information and the other AUs untouched.Owing to the  dataset BP4D with poor diversity and unbalanced labels distribution, the generated results we trained is not as well as in their paper. However, ours model can be trained to generate the better results with the same dataset.

\vspace{-2mm}

\vspace{-2mm}
\subsubsection{Synthetic Results of AU combinations:}
\begin{figure}[!htb]
\centering
\includegraphics[width=0.6\textwidth]{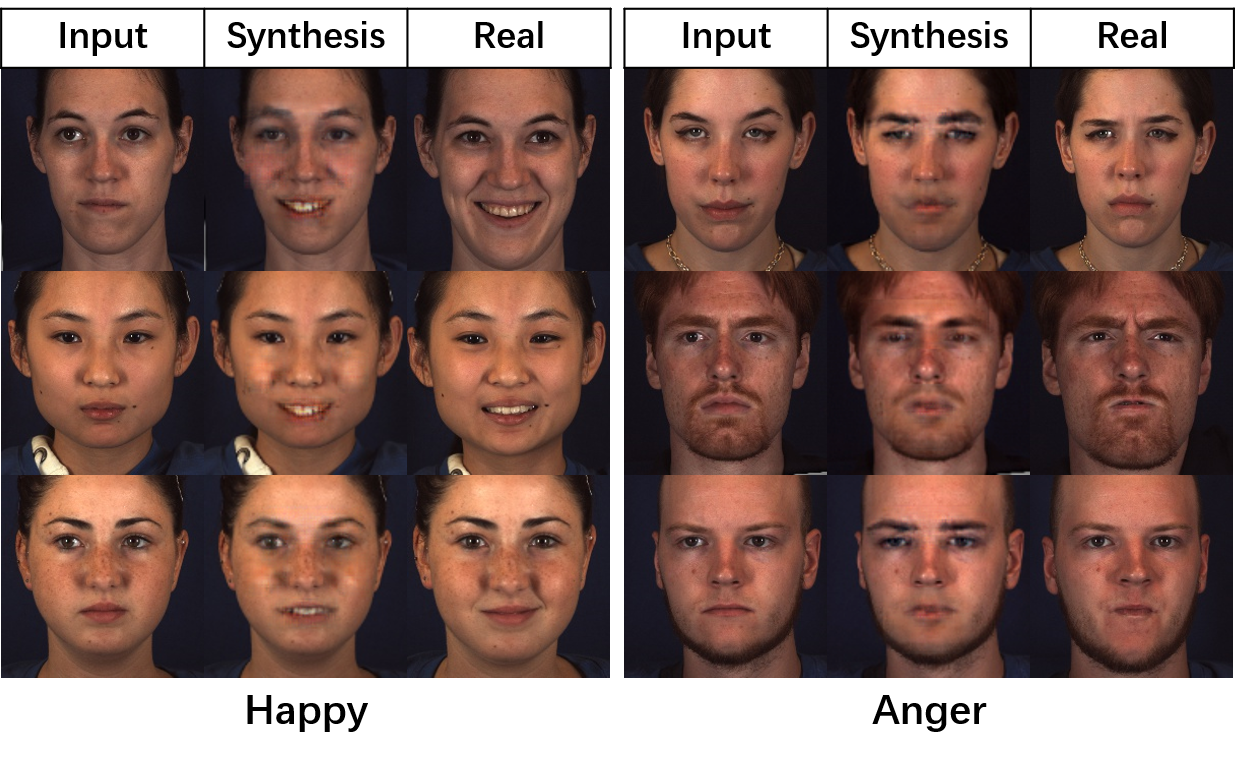}
\caption{Synthetic Results of Common AU combinations.} \label{fig5} 
\end{figure}

Our LAC-GAN can synthesize spontaneous facial expression incorporated by multiple AUs, such as happy and angry. Fig.~\ref{fig5} shows the generated happy (AU2, AU6, AU7 and AU12) and angry (AU4 and AU23), which are substantially similar to the ground-truth expressions. The slight difference is probably caused by the different intensity of the presence of AUs. 

\vspace{-2mm}
\vspace{-2mm}

\vspace{-5mm}
\subsection{Quantitative Analysis of the Synthetic Results}
\vspace{-2mm}
To assess whether the synthetic images are capable of being accurately detected by the AU detection methods, we evaluate our synthetic images with state-of-the-art AU detection models. We firstly generated the augmentation dataset by our synthesis model of which the person's subject and corresponding AUs labels distribution is same as the ground-truth test dataset .Afterwards, the real images and synthesis images are estimated by two AU detectors respectively which are OpenFace AU classifier ~\cite{cmusatyalab@OpenFace} and JAA-Net~\cite{Shao2018JAA}.These two models are demonstrated state-of-the-art results in the task of AU  detection.
F1 score and accuracy are adopted as metrics. In the following section, all the results are simplified without \%.


\begin{table*}[htb]
\centering
\caption{Quantitative Analysis Of Our Synthetic Results. }\label{tab2}
\begin{tabular}{|c|c|c|c|c|c|c|c|c|c|c|c|c|}
\hline
\multirow{3}{*}{AU} & \multicolumn{6}{|c|}{OpenFace}& \multicolumn{6}{|c|}{JAA-Net} \\ \cline{2-13}
&\multicolumn{2}{|c|}{Real}& \multicolumn{2}{|c|}{Synthesis}&\multicolumn{2}{|c|}{Gaps}& \multicolumn{2}{|c|}{Real}& \multicolumn{2}{|c|}{Synthesis}&\multicolumn{2}{|c|}{Gaps} \\ \cline{2-13}
& F1 & Acc & F1 & Acc & F1 & Acc & F1 & Acc & F1 & Acc & F1 & Acc\\ \hline
AU1 & 64.1 & 76.1 & 75.8 & 88.7 & -11.7 & -12.6 & 46.2 & 69.7 & 59.5 & 76.5 & -13.3 & -6.8\\ 
AU2 & 44.3 & 66.2 & 33.0 & 51.3 & 11.3 & 14.9 & 48.7 & 78.4 & 33.7 & 67.1 & 15 & 11.3\\ 
AU4 & 71.9 & 87.4 & 67.7 & 77.6 & 4.2 & 9.8 & 56.4 & 84.7 & 52.6 & 78.7 & 3.8 & 6\\ 
AU6 & 84.3 & 83.5 & 73.1 & 75.2 & 11.2 & 8.3 & 80.4 & 80.2 & 72.3 & 75.9 & 8.1 & 4.3\\ 
AU7 & 77.6 & 76.8 & 69.5 & 69.4 & 8.1 & 7.4 & 73.5 & 72.9 & 67.2 & 65.8 & 6.3 & 7.1 \\ 
AU10 & 88.2 & 85.0 & 76.8 & 72.3 & 11.4 & 12.7 & 84.9 & 80.4 & 74.3 & 68.5 & 10.6 & 11.9\\ 
AU12 & 73.6 & 68.7 & 72.8 & 78.0 & 0.8 & -9.3 & 88.4 & 85.7 & 83.6 & 80.1 & 4.8 & 5.6\\ 
AU14 & 91.0 & 89.2 & 85.1 & 84.8 & 5.9 & 4.4 & 59.4 & 60.2 & 63.7 & 65.1 & -4.3 & -4.9\\ 
AU15 & 44.3 & 85.7 & 35.7 & 76.7 & 6.8 & 9 & 45.6 & 84.7 & 34.1 & 76.4 & 11.5 & 8.3\\ 
AU17 & 69.6 & 78.8 & 40.4 & 51.2 & 29.3 & 27.6 & 61.1 & 73.3 & 35.8 & 54.7 & 25.3 & 18.6\\ 
AU23 & 53.6 & 80.3 & 47.5 & 74.9 & 6.1 & 5.4 & 36.2 & 81.0 & 31.7 & 74.9 & 4.5 & 6.1\\ 
AU24 & - & - & - & - & - & - & 37.4 & 83.9 & 30.6 & 76.2 & 6.8 & - 7.7\\ \hline
Avg & 69.3 & 79.8 & 61.7 & 72.7 & 7.6 & 7.1 & 59.9 & 77.9 & 53.3 & 71.7 & 6.6 & 6.3\\ \hline
\end{tabular}
\end{table*}

Table~\ref{tab2} shows the quantitative analysis results of synthesis images augmentation. 'Gaps' is regarded as the different value between the detection results of 'Real' dataset   and 'Synthesis'dataset. '$-$' means there are no results of AU24 with OpenFace classifier.Specifically, the results of synthesis images approaches to the real images,  between which the gaps is approximately $6\% \sim 7\% $ with average metrics which means our synthesis images are highly homologous with the real images on AU detection level. While JAA-Net used as the detector, the gaps between Real and Synthesis is less than while OpenFace is used as the detector, on account of the local AU attention localization JAA-Net adopts similar with our model. AU1, AU4, AU12, AU14, AU23 generated by our model is demonstrably close to the the ground truth images with both detectors, especially for AU1 and AU14. However,the performance of generated face on AU17 is poor, which accounts the occurrence of AU17 is lower than the other AUs in the dataset. Substantially, the images generated by our model is effective to be detected by the state-of-the-art AU detection models.

\vspace{-2mm}
\section{Conclusion}
\vspace{-2mm}
In this paper, we presents LAC-GAN for facial action units synthesis, which is incorporated into the photo-realistic facial expressions, without destroying the other facial information except the AU. The generator can manipulate the AU in different status, such as transferring from absence to presence. The local discriminator and AU classifier in CARGs guarantees the consistency between the desired faces and the corresponding target AU. The AU regional localization is proposed to combine all the AU regions to an integral face with keep the background untouched. We further conducted extensive experimental quantitative analysis to evaluate our synthesis model. Our future work will explore how to apply LAC-GAN to other larger and more unconstrained facial expression dataset with intensity-level generation.

\vspace{-3mm}
\section*{Acknowledgements}
\vspace{-3mm}
This work is supported by the National Natural Science Foundation of China under Grants of 41806116 and 61503277. We gratefully acknowledge the support of NVIDIA Corporation with the donation of the Titan V GPU used for this research.
%
%
%
%
\vspace{-2mm}
\bibliographystyle{splncs04}
\bibliography{bibtex-example.bib}
\end{document}